\documentclass[numbers]{article}

\usepackage[preprint]{neurips_2025}

\usepackage[utf8]{inputenc} 
\usepackage[T1]{fontenc}    
\usepackage{hyperref}       
\usepackage{url}            
\usepackage{booktabs}       
\usepackage{amsfonts}       
\usepackage{nicefrac}       
\usepackage{microtype}      
\usepackage{xcolor}         
\usepackage{textgreek}
\usepackage{amsmath}
\usepackage{graphicx}
\usepackage[utf8]{inputenc}
\usepackage{newunicodechar}

\title{Cross-Temporal Attention Fusion (CTAF) for Multimodal Physiological Signals in Self-Supervised Learning}

\author{
  Arian Khorasani \\
  {Department of Information Technologies, HEC Montréal} \\
  {Mila-Quebec Artificial Intelligence Institute} \\
  \texttt{Arian.Khorasani@mila.quebec}
  \And
  Théophile Demazure \\
  {Department of Information Technologies, HEC Montréal} \\
  \texttt{theophile.demazure@hec.ca} \\
}

\begin{document}

\maketitle

\begin{abstract}

We study multimodal affect modeling when EEG and peripheral physiology are asynchronous, which most fusion methods ignore or handle with costly warping. We propose Cross-Temporal Attention Fusion (CTAF), a self-supervised module that learns soft bidirectional alignments between modalities and builds a robust clip embedding using time-aware cross attention, a lightweight fusion gate, and alignment-regularized contrastive objectives with optional weak supervision. On the K-EmoCon dataset, under leave-one-out cross-validation evaluation, CTAF yields higher cosine margins for matched pairs and better cross-modal token retrieval within one second, and it is competitive with the baseline on three-bin accuracy and macro-F1 while using few labels. Our contributions are a time-aware fusion mechanism that directly models correspondence, an alignment-driven self-supervised objective tailored to EEG and physiology, and an evaluation protocol that measures alignment quality itself. Our approach accounts for the coupling between the central and autonomic nervous systems in psychophysiological time series. These results indicate that CTAF is a strong step toward label-efficient, generalizable EEG-peripheral fusion under temporal asynchrony.

\end{abstract}

\section{Introduction}

Predicting latent psychophysiological constructs from multimodal biosignals is hard because modalities operate on different time scales and are not perfectly synchronous. Electroencephalography (EEG), electrodermal activity (EDA), blood-volume pulse (BVP), and electrocardiography (ECG) offer complementary views of central and autonomic systems, yet their latencies and dynamics differ in ways that matter. Labeled data are also scarce while models must generalize across tasks and subjects \citep{aristimunha2025eegfoundationchallengecrosstask}. Recent surveys recommend handling imperfect synchronization, supporting cross-modal retrieval, and enforcing representation invariance to improve robustness in these settings \citep{jiang2025multimodaltimeseriesanalysis, 8269806}.

Self-supervised learning provides a route to exploit abundant unlabeled sensor data while encouraging invariance to nuisance variability. Contrastive and redundancy-reduction objectives promote stability to sensor noise and subject differences \citep{jiang2025multimodaltimeseriesanalysis, chen2020simpleframeworkcontrastivelearning, grill2020bootstraplatentnewapproach, zbontar2021barlowtwinsselfsupervisedlearning, bardes2022vicregvarianceinvariancecovarianceregularizationselfsupervised}. Time-series SSL further addresses local stochasticity and temporal warping through context and view consistency \citep{yue2022ts2vecuniversalrepresentationtime, sermanet2018timecontrastivenetworksselfsupervisedlearning}. In physiology, cross-lead and cross-timestamp pretraining has improved ECG tasks without labels, underscoring the promise of SSL in this domain \citep{kiyasseh2021clocscontrastivelearningcardiac}.

Building on this foundation, we introduce Cross-Temporal Attention Fusion (CTAF), a self-supervised alignment-and-fusion module for synchronized yet imperfectly aligned multimodal physiological time series. CTAF learns soft cross-temporal correspondences between modalities instead of assuming frame-level synchrony, and it couples these alignments with contrastive self-supervision and redundancy-reduction to avoid collapse \citep{sermanet2018timecontrastivenetworksselfsupervisedlearning, wang2019learningcorrespondencecycleconsistencytime, chen2020simpleframeworkcontrastivelearning, radford2021learningtransferablevisualmodels, bardes2022vicregvarianceinvariancecovarianceregularizationselfsupervised, zbontar2021barlowtwinsselfsupervisedlearning}. The design is mask-aware for missing tokens and label-efficient, and the same encoders can be probed with supervised heads when labels are available without architectural changes \citep{bardes2022vicregvarianceinvariancecovarianceregularizationselfsupervised}.

We evaluate CTAF under subject-wise leave-one-out cross-validation and compare against strong hypercomplex baselines \citep{10193329, lopez2024hierarchicalhypercomplexnetworkmultimodal, lopez2024phemonetmultimodalnetworkphysiological}. Empirically, CTAF improves cross-modal retrieval under a temporal tolerance and increases the separation between matched and mismatched clip embeddings, and it is competitive on discretized recognition while using the same preprocessed inputs and splits. Our contributions are a timing-aware self-supervised fusion module that explicitly handles cross-modal lags, an evaluation protocol centered on alignment and cross-subject generalization, and a reproducible baseline suite for multimodal time-series learning \citep{jiang2025multimodaltimeseriesanalysis}

\section{Related Works}

\textbf{Self-supervised representation learning for time series}. In self-supervised learning (SSL), models are trained on tasks where the supervision signal is derived from the input data itself, without requiring labels. Self-supervised learning for time series has advanced rapidly in recent years.  Contrastive predictive coding \citep{oord2019representationlearningcontrastivepredictive} proposed SSL for sequences. Subsequent work refined this with temporal neighborhoods coding  \citep{tonekaboni2021unsupervisedrepresentationlearningtime}, hierarchical and temporal contrasts \citep{yue2022ts2vecuniversalrepresentationtime}, and temporal/contextual contrasting \citep{eldele2021timeseriesrepresentationlearningtemporal}. CTAF builds on this line of research, but selects positive pairs using learned cross-temporal alignment rather than strictly synchronous counterparts. 
\vspace{1pt}

\textbf{Brain and Body neurophysiological responses}. The central and the autonomic nervous systems are tightly coupled but operate on different time scales. For example, EEG responses occur within tens of milliseconds, phasic EDA responses time by several seconds, and HRV changes evolve more gradually. Despite these differences, the brain and body function in a coordinated and adaptive manner \citep{BENEDEK201080, THAYER2000201}. Our approach  accounts for this brain–body asynchrony through time-aware fusion.
\vspace{1pt}

\textbf{Cross-modal fusion and alignment}. Cross-modal fusion and alignment are central challenges in multimodal learning. Classical multi-view correlation approaches, such as Deep Canonical Correlation Analysis \citep{pmlr-v28-andrew13}, and multimodal Transformers have addressed unaligned input streams. For example, MulT applies directional cross-modal attention to handle unaligned sequences \citep{tsai-etal-2019-multimodal}. CTAF differs by explicitly learning lag distributions via cross-temporal attention, and leveraging these learned alignments to guide the self-supervised learning objective. 
\vspace{1pt}

\textbf{Differentiable alignment and cycle consistency}. Soft-DTW offers a differentiable time-warping loss for sequences \citep{cuturi2018softdtwdifferentiablelossfunction}; temporal cycle-consistency methods in video have been used to learn correspondences without labels \citep{dwibedi2019temporalcycleconsistencylearning,wang2019learningcorrespondencecycleconsistencytime}. CTAF borrows the same intuition, imposing bidirectional EEG $\leftrightarrow$ peripheral measures consistency to regularize alignment maps. 
\vspace{1pt}

\textbf{Multimodal electrophysiology with contrastive/attention fusion}. Recent work has applied SSL and attention mechanisms to EEG in combination with EOG, HR, or EDA signals for sleep staging or emotion recognition, typically assuming synchronicity and fixed windows \citep{10.1145/3608164.3608185, guo2023emotionrecognitionbasedmultimodal}. CTAF learns soft, time-shifted correspondences to select cross-modal positives, producing generalizable embeddings that reflect heterogeneous, but relevant, temporal dynamics across modalities.
\vspace{1pt}

\textbf{Multimodal affect datasets (EEG + peripheral measures)}. Community benchmarks pairing EEG with autonomic/peripheral signals (BVP/ECG/EDA), notably DEAP, AMIGOS, WESAD, and K-EmoCon, established realistic cross-modal timing issues and subject variability that motivate time-aware fusion and evaluation\citep{16d2153c76974234a7af4ed017b3bb68, Park_2020, 8554112, 10.1145/3242969.3242985}. In this work, we use K-EmoCon as our primary dataset.
\vspace{1pt}

\section{Cross-Temporal Attention Fusion (CTAF)}

CTAF is a self-supervised fusion module for EEG and peripheral physiology that learns time-aware cross-temporal correspondences and a joint clip representation. For notational consistency we use superscripts $(e)$ for EEG and $(p)$ for peripheral physiology throughout. Let

$$
X^{(e)}\in\mathbb{R}^{S_e\times D_e},\qquad 
X^{(p)}\in\mathbb{R}^{S_p\times D_p}
$$

denote per-bin feature sequences with timestamp vectors

$$
t^{(e)}\in\mathbb{R}^{S_e},\qquad 
t^{(p)}\in\mathbb{R}^{S_p},
$$

and validity masks

$$
m^{(e)}\in\{0,1\}^{S_e},\qquad 
m^{(p)}\in\{0,1\}^{S_p}.
$$

CTAF outputs (i) soft cross-temporal alignment matrices $A_{e\to p}\in\mathbb{R}^{S_e\times S_p}$ and $A_{p\to e}\in\mathbb{R}^{S_p\times S_e}$ that capture token-level correspondences despite unknown lags, and (ii) a robust clip-level embedding $z_f\in\mathbb{R}^d$ (along with modality projections $z^{(e)},z^{(p)}\in\mathbb{R}^d$) suitable for subject-generalizable downstream use.

\subsection{Preliminaries and Problem Setup}

We consider paired sequences $\big(X^{(e)}, t^{(e)}, m^{(e)}\big)$ and $\big(X^{(p)}, t^{(p)}, m^{(p)}\big)$ from the same window, sampled at different effective rates with possible dropouts. CTAF has two goals: (1) learn modality-invariant clip embeddings $z^{(e)}$ and $z^{(p)}$ that coincide for the same window, and (2) learn token-level correspondences between $X^{(e)}$ and $X^{(p)}$ that allow variable, sample-dependent lags.

Misalignment is inherent in EEG–autonomic coupling, so CTAF models soft, time-varying correspondences rather than assuming synchrony, learns token- and clip-level links jointly, and forms time-aware positives for contrastive learning instead of enforcing simultaneity. The design is label-efficient and mask-aware, respecting $m^{(e)}$ and $m^{(p)}$ for missing tokens, and it avoids heavy alignment path solvers, which keeps training scalable under realistic cross-modal timing noise.

\subsection{CTAF Architecture}

\textbf{Modality encoders.} Each stream is encoded with a lightweight Conv-Transformer that is both time-aware and mask-aware. With inputs $X^{(e)}\in\mathbb{R}^{S_e\times D_e}$, $X^{(p)}\in\mathbb{R}^{S_p\times D_p}$, timestamps $t^{(e)},t^{(p)}$ and validity masks $m^{(e)},m^{(p)}$,

$$
H^{(e)}=\mathrm{Enc}_e\!\big(X^{(e)},\,\phi(t^{(e)});\;\text{kpm}=\neg m^{(e)}\big),\qquad
H^{(p)}=\mathrm{Enc}_p\!\big(X^{(p)},\,\phi(t^{(p)});\;\text{kpm}=\neg m^{(p)}\big),
$$

where $\phi(\cdot)$ are sinusoidal time features (added or concatenated), and “kpm” is the key-padding mask used to prevent attention to invalid tokens. Encoders output $H^{(e)}\in\mathbb{R}^{S_e\times d}$, $H^{(p)}\in\mathbb{R}^{S_p\times d}$.

\textbf{Bidirectional cross-temporal attention.} To expose cross-modal lags, we apply multi-head attention in both directions: 

\begin{equation}
\begin{aligned}
\tilde H^{(e)} &= \mathrm{MHA}\!\left(Q{=}H^{(e)},\,K{=}H^{(p)},\,V{=}H^{(p)};\ \text{kpm}=\neg m^{(p)}\right),\\
\tilde H^{(p)} &= \mathrm{MHA}\!\left(Q{=}H^{(p)},\,K{=}H^{(e)},\,V{=}H^{(e)};\ \text{kpm}=\neg m^{(e)}\right).
\end{aligned}
\end{equation}

The resulting attention maps are the soft cross-temporal alignments $A_{e\to p}=\mathrm{softmax}(Q_eK_p^\top/\sqrt d)$ and $A_{p\to e}$, with invalid keys masked to $-\infty$.

\textbf{Global summaries, fusion gate, and token pooling. } We use two explicit pooling operators:

(1) Masked mean pooling for global summaries:
$$
\mathrm{Mean}_m(H)=\frac{\sum_i m_i H_i}{\sum_i m_i+\varepsilon},\qquad
z^{(e)}=\mathrm{Mean}_{m^{(e)}}(\tilde H^{(e)}),\;\; z^{(p)}=\mathrm{Mean}_{m^{(p)}}(\tilde H^{(p)}).
$$

(2) Masked attention pooling for fused tokens: we first form $T=\tfrac12(\tilde H^{(e)}+\tilde H^{(p)})$ with union mask $m^{(\cup)}=m^{(e)}\lor m^{(p)}$. With a learned query $q\in\mathbb{R}^d$,

$$
\alpha_i \propto \exp\!\Big(\tfrac{q^\top T_i}{\sqrt d}\Big)\ \text{for valid }i,\quad
\alpha_i{=}0\ \text{if }m^{(\cup)}_i{=}0,\quad
z_{\text{tok}}=\sum_i \alpha_i T_i.
$$

A lightweight fusion gate mixes the modality summaries in a data-dependent way:

$$
g=\sigma\!\big(\mathrm{MLP}([z^{(e)}\!\parallel z^{(p)}])\big),\qquad
z_{\text{gate}}=g\odot z^{(e)}+(1-g)\odot z^{(p)}.
$$

The final clip embedding is

$$
z_f=\tfrac12\big(z_{\text{gate}}+z_{\text{tok}}\big),
$$

i.e., the fusion gate combines global summaries, and an attention layer pools over fused tokens to yield the clip-level representation.

\textbf{Projection heads. } Three MLP heads map to contrastive space (with $\ell_2$ normalization): 

$$
p^{(e)}=\mathrm{Proj}_e(z^{(e)}),\quad p^{(p)}=\mathrm{Proj}_p(z^{(p)}),\quad p^{(f)}=\mathrm{Proj}_f(z_f).
$$

Token-level projections are used when computing alignment/retrieval losses.

Masks are enforced at every sequence-consuming step: (i) in encoders and cross-attention via key-padding masks; (ii) in global summaries via the masked mean; (iii) in attention pooling by zeroing logits of invalid positions before softmax; and (iv) in token-level losses, which operate only on valid indices (or on the union mask for fused tokens). This makes CTAF permutation-invariant over valid tokens, explicitly time-aware via $\phi(t)$, and robust to missing data.

\subsection{Learning Objectives and Training Strategy}

CTAF couples complementary terms that bind the modalities semantically, learn time-aware token matches, and stabilize the representation.

\textbf{Clip-Level Cross-Modal Contrast (InfoNCE \citep{chen2020simpleframeworkcontrastivelearning}).} Let $z_e,z_p\in\mathbb{R}^d$ be $L^2$-normalized clip projections for EEG and physiology. With temperature $T$ and batch index $i$,

$$
\mathcal{L}_{\text{con}}
=\tfrac12\!\sum_i\!\Big[
-\log\tfrac{\exp(\langle z^i_e,z^i_p\rangle/T)}{\sum_j \exp(\langle z^i_e,z^j_p\rangle/T)}
-\log\tfrac{\exp(\langle z^i_p,z^i_e\rangle/T)}{\sum_j \exp(\langle z^i_p,z^j_e\rangle/T)}
\Big].
$$

\textbf{Soft Cross-Temporal Alignment.} Project token sequences to $\tilde H^{\text{eeg}},\tilde H^{\text{phys}}$ and normalize tokens. For sample $i$, token similarity $S^i_{su}=\langle \tilde h^{\text{eeg}}_{s},\tilde h^{\text{phys}}_{u}\rangle/T$. Define soft targets by a Gaussian over time deltas,

$$
W^i_{su}\propto \exp\!\Big(-\tfrac{(t^{\text{eeg}}_s - t^{\text{phys}}_u)^2}{2\sigma^2}\Big),\quad
\sum_u W^i_{su}=1.
$$

We minimize row- and column-wise cross-entropies:

$$
\mathcal{L}_{\text{align}}
= \tfrac12\!\sum_i \Big[
-\!\sum_s \sum_u W^i_{su}\,\log \operatorname{softmax}_u(S^i_{su})
-\!\sum_u \sum_s \hat W^i_{us}\,\log \operatorname{softmax}_s(S^i_{su})
\Big],
$$

with $\hat W$ the column-normalized transpose, encouraging near-synchronous matches while tolerating realistic lags.

\textbf{Fusion.} Let $z_f$ be the fused clip projection. We tether it to the modalities with

$$
\mathcal{L}_{\text{fuse}} = \big\| z_f - \tfrac12(z_e+z_p) \big\|_2^2.
$$

\textbf{VICReg-Style Stabilization.} On pre-normalized projections $p_e,p_p$, we use invariance, variance, and covariance penalties:

$$
\mathcal{L}_{\text{inv}}=\|p_e-p_p\|_2^2,\quad
\mathcal{L}_{\text{var}}=\tfrac12\!\sum_{m\in\{e,p\}}\!\!\operatorname{mean}\big[\max(0,1-\operatorname{std}(p_m))\big],
$$

$$
\mathcal{L}_{\text{cov}}=\tfrac12\!\sum_{m\in\{e,p\}}\!\!\operatorname{mean}\big[\operatorname{offdiag}(\operatorname{Cov}(p_m))^2\big].
$$

\textbf{Within-Modality View Contrast.} Two stochastic time views per stream (warp/jitter/mask) yield $\mathcal{L}_{\text{view}}$ via symmetric InfoNCE \citep{chen2020simpleframeworkcontrastivelearning} within each modality, improving robustness to sampling and segmentation.

\textbf{Consistency Under Time Jitter.} Re-encode the same window with different time jitters; penalize drift of $z_f$:

$$
\mathcal{L}_{\text{cons}}=\| z_f^{(1)}-z_f^{(2)}\|_2^2.
$$

\textbf{Auxiliary Supervision.} When window-level labels $y\in\mathbb{R}^2$ (arousal, valence) are available, a lightweight MLP head on the fused clip representation $z_f$ predicts $\hat y$. We train this head with a small weight using z-scored targets

$$
y_z \;=\; (y-\mu)\oslash\sigma,\qquad \mu,\sigma\in\mathbb{R}^2
$$

where $\mu$ and $\sigma$ are the per-dimension mean and standard deviation computed on the training windows of the current fold, and $\oslash$ denotes element-wise division. The loss is

$$
\mathcal{L}_{\text{sup}} \;=\; \lVert \hat y - y_z \rVert_2^2,
$$

applied only to windows with valid labels; this term is auxiliary and does not drive the main alignment objective.

\textbf{Total Objective.} With scheduled weights $\beta_t,\alpha_f,\lambda_{\text{inv}},\lambda_{\text{var}},\lambda_{\text{cov}},\lambda_{\text{view}},\lambda_{\text{cons}},\lambda_{\text{sup}}$,

$$
\mathcal{L}_{\text{CTAF}}
= \mathcal{L}_{\text{con}}
+ \beta_t\,\mathcal{L}_{\text{align}}
+ \alpha_f\,\mathcal{L}_{\text{fuse}}
+ \lambda_{\text{inv}}\mathcal{L}_{\text{inv}}
+ \lambda_{\text{var}}\mathcal{L}_{\text{var}}
+ \lambda_{\text{cov}}\mathcal{L}_{\text{cov}}
+ \lambda_{\text{view}}\mathcal{L}_{\text{view}}
+ \lambda_{\text{cons}}\mathcal{L}_{\text{cons}}
+ \lambda_{\text{sup}}\mathcal{L}_{\text{sup}}.
$$

\subsection{Other Design Choices}

We use a short curriculum: the time-jitter amplitude and the weight $\beta_t$ on $\mathcal{L}_{\text{align}}$ increase over training. The model first binds modalities at the clip level, then sharpens token-level lag handling. Robustness to missing sensors is encouraged via modality dropout—randomly zeroing an entire stream—together with mask-aware pooling.
\vspace{1pt}

CTAF is trained largely self-supervised to learn a time-aware shared latent. Dual Transformer encoders with sinusoidal time features produce modality sequences; bi-directional cross-attention exchanges information; a lightweight fusion gate mixes global EEG and physiology summaries; and attention pooling over fused tokens yields a clip embedding. The objective combines bidirectional InfoNCE, a soft temporal-alignment loss that tolerates realistic offsets, a VICReg regularizer (invariance/variance/covariance), within-modality view contrast via jitter/warp perturbations, and a consistency term for time-augmented encodings. A small supervised head provides a weak regression signal for arousal/valence. Architecturally we rely on standard self-attention blocks \citep{vaswani2023attentionneed} and modern self-supervised components \citep{oord2019representationlearningcontrastivepredictive, chen2020simpleframeworkcontrastivelearning, bardes2022vicregvarianceinvariancecovarianceregularizationselfsupervised} adapted to cross-modal, temporally misaligned data.  
\vspace{1pt}

\section{Empirical Evaluation}

We evaluate CTAF on K-EmoCon, which provides continuous 5-second arousal–valence annotations with peripheral physiology from Empatica E4 and EEG from NeuroSky \citep{Park_2020}. All methods use identical fixed-length windows from our preprocessing, where we bin each stream into equal-width tokens, compute a 10-channel EEG feature vector and a 4-channel physiology vector covering BVP, EDA, skin temperature, and heart rate, attach the window-level self-report, and apply subject-wise per-channel normalization. We further require a minimum joint coverage of valid EEG and physiology tokens to limit label leakage, and we batch windows without altering their content.
\vspace{1pt}

We evaluate with subject-level leave-one-out cross-validation (LOOCV). For each participant we train on the remaining $ N-1 $ with fixed hyperparameters, choose the CTAF checkpoint that minimize the validation self-supervised objective, and then test on the held-out subject. Our main alignment metrics are clip-level cosine similarity comparing matched to mismatched EEG-physiology pairs and token-level cross-modal retrieval accuracy within a tolerance $\tau$ seconds in both EEG→phys and phys→EEG directions, computed both with and without time features to assess learning beyond explicit timing. For interpretability we also report accuracy and macro-F1 on three pre-registered bins for arousal and valence.
\vspace{1pt}

For the baseline, we compare our CTAF against HyperFuseNet, a parameter-efficient hypercomplex fusion network that replaces large real-valued layers with parameterized hypercomplex multiplication and quaternionic operators to exploit inter-channel structure \citep{lopez2024hierarchicalhypercomplexnetworkmultimodal, pmlr-v28-andrew13, 10193329}. We re-implement the HyperFuseNet head and its modality branches, adapt inputs to our windowed layout, and train it in a supervised manner using the official schedule. Following prior work, we report three-bin accuracy and macro-F1 for HyperFuseNet, which are computed on the same windows and LOOCV splits. 
\vspace{1pt} 

\section{Results}

\textbf{Baseline comparison.} Under identical preprocessing and LOOCV splits, CTAF outperforms HyperFuseNet on three-bin arousal/valence recognition. Table \ref{tab:baseline_comparison} shows higher accuracy and macro-F1 for CTAF, indicating that its time-aware embeddings support stronger discretized recognition.


\begin{table}[t]
\centering
\caption{Baseline comparison (3-bin classification). Values are reported as mean $\pm$ 95\% confidence interval across LOOCV participants.}
\label{tab:baseline_comparison}
\begin{tabular}{lcc}
\toprule
Metric & \textbf{CTAF (ours)} & \textbf{HyperFuseNet} \\
\midrule
Accuracy & \textbf{0.62} $\pm$ 0.04 & 0.58 $\pm$ 0.05 \\
Macro-F1 & \textbf{0.61} $\pm$ 0.04 & 0.57 $\pm$ 0.05 \\
\bottomrule
\end{tabular}
\end{table}

\begin{table}[t]
\centering
\caption{Cross-modal alignment under LOOCV on K-EmoCon ($\tau{=}1$\,s). Macro means and 95\% CIs are reported across held-out participants. Metrics: “cos\_pos” (matched EEG–physiology clips), “cos\_neg” (mismatched clips), and token retrieval rates “retr@$\,\tau$ e2p” (EEG$\rightarrow$physiology) and “retr@$\,\tau$ p2e” (physiology$\rightarrow$EEG). The lower panel shows absolute improvements when enabling time features (WITH time minus NO time).}
\label{tab:alignment_all}
\small
\begin{tabular}{lcc}
\toprule
\multicolumn{3}{c}{\textbf{Macro means [95\% CI]}}\\
\midrule
Metric & WITH time & NO time \\
\midrule
cos\_pos & 0.240 \,[0.219,\,0.259] & 0.051 \,[0.033,\,0.067] \\
cos\_neg & 0.004 \,[-0.007,\,0.015] & 0.003 \,[-0.008,\,0.014] \\
retr@$\tau$ e2p & 0.350 \,[0.334,\,0.366] & 0.212 \,[0.207,\,0.220] \\
retr@$\tau$ p2e & 0.265 \,[0.247,\,0.284] & 0.208 \,[0.206,\,0.210] \\
\bottomrule
\end{tabular}

\vspace{0.6em}

\begin{tabular}{lcc}
\toprule
\multicolumn{3}{c}{\textbf{WITH time $-$ NO time (absolute gain)}}\\
\midrule
Metric & $\Delta$ mean & 95\% CI \\
\midrule
cos\_pos & +0.189 & [\,0.163,\,0.215\,] \\
cos\_neg & +0.001 & [\,-0.010,\,0.012\,] \\
retr@$\tau$ e2p & +0.138 & [\,0.119,\,0.156\,] \\
retr@$\tau$ p2e & +0.057 & [\,0.038,\,0.075\,] \\
\bottomrule
\end{tabular}
\end{table}

\textbf{Cross-modal alignment.} Under LOOCV, CTAF consistently learns time-aware EEG–physiology correspondence. In Table \ref{tab:alignment_all} Panel A, matched clips cluster well above mismatches (cosine $0.240$ [$0.219$, $0.259$] vs $0.051$ [$0.033$, $0.067$]), and Figure $1$ shows this separation across participants. Figure $2$ shows most points above the diagonal, indicating consistent gains across subjects. The effect persists when absolute time is removed (Table \ref{tab:alignment_all}, Figure \ref{fig:cos-scatter}), with a median per-participant gain of $0.182$ (IQR [$0.141$, $0.224$]), indicating the model captures correspondence beyond explicit timestamps. 

\begin{figure}
    \centering
    \includegraphics[width=0.7\linewidth]{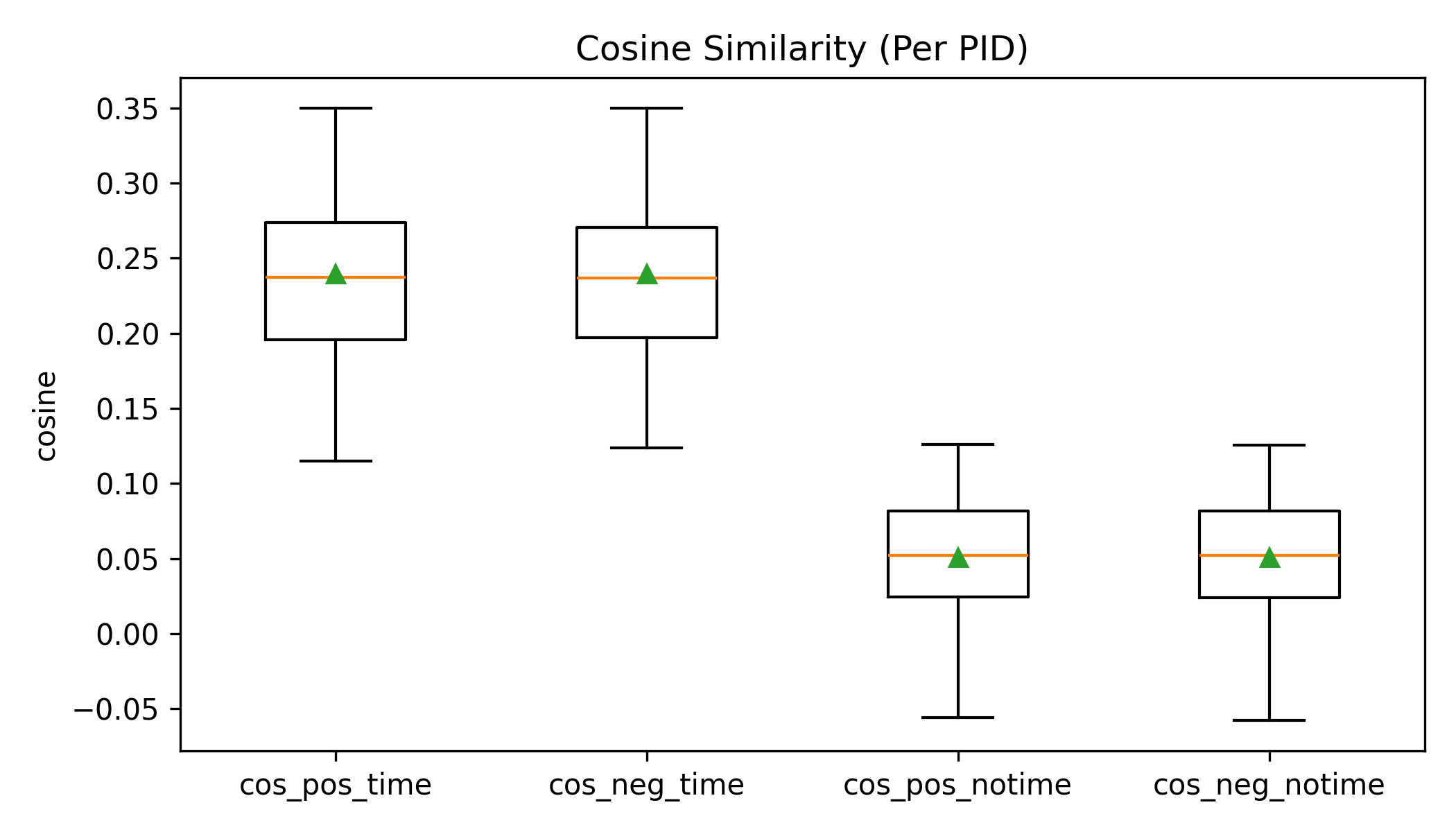}
    \caption{Clip-level cosine similarity distributions for matched (cos\_pos) vs. mismatched (cos\_neg) EEG–physiology pairs, aggregated over LOOCV folds. Larger separation indicates stronger cross-modal alignment; boxes show median and IQR across participants (PIDs).}
    \label{fig:cos-box}
\end{figure}

\begin{figure}
    \centering
    \includegraphics[width=0.5\linewidth]{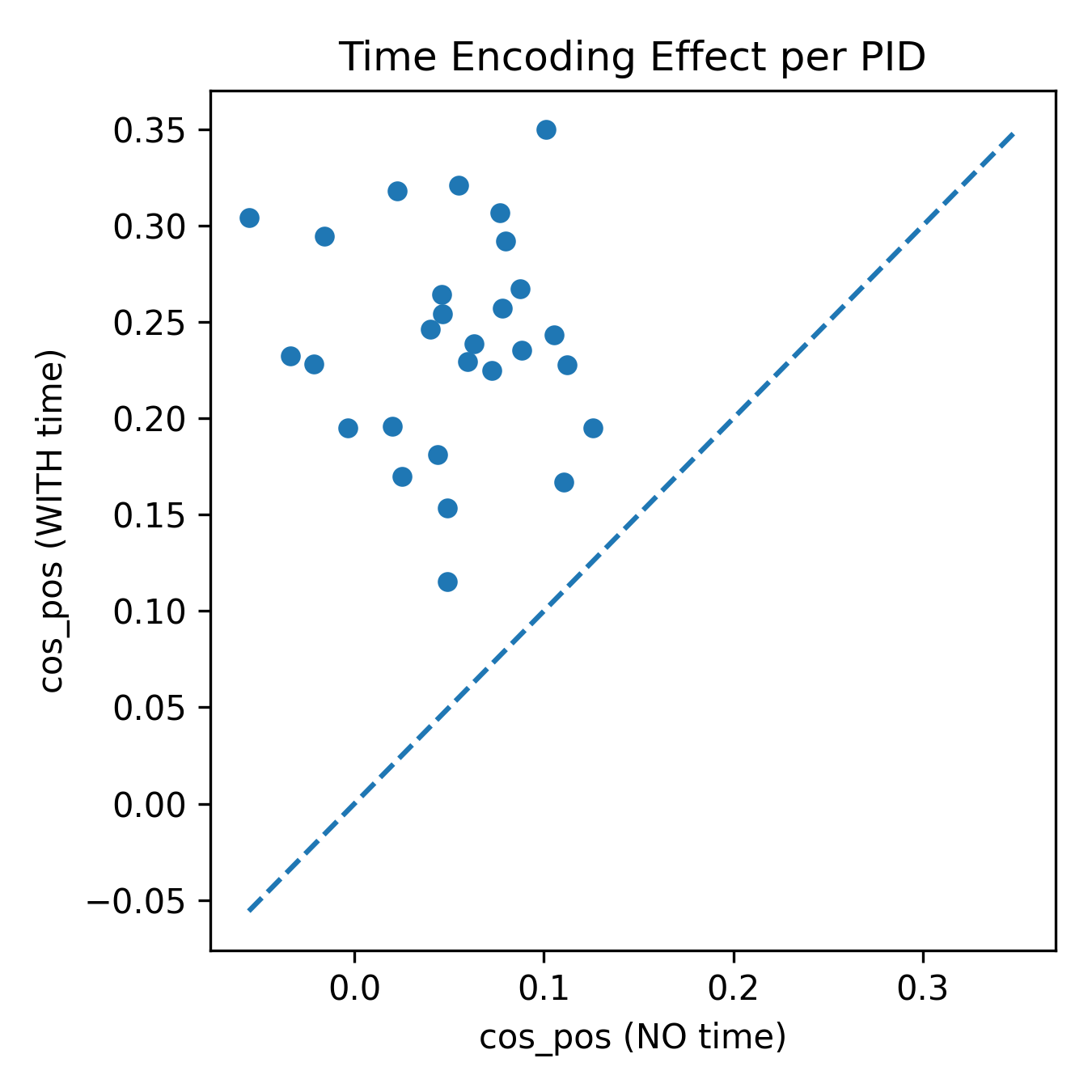}
    \caption{Per-participant (PID) matched cosine (cos\_pos) \emph{WITH time (CTAF)} vs. \emph{NO time} ablation. Points above the diagonal indicate improved alignment when using CTAF’s time-aware fusion.}
    \label{fig:cos-scatter}
\end{figure}

\textbf{Clip-level cosine.} Figure \ref{fig:cos-box} summarizes per-participant cosine similarity for matched EEG-physiology clips versus mismatched pairs. With time encodings enabled, matched pairs trend higher; Figure 2 shows per-subject with time in the y-axis against no time in the x-axis, and most points lie above the diagonal, indicating that CTAF extracts stronger correspondences when it can exploit temporal features. 

\begin{figure}
    \centering
    \includegraphics[width=0.6\linewidth]{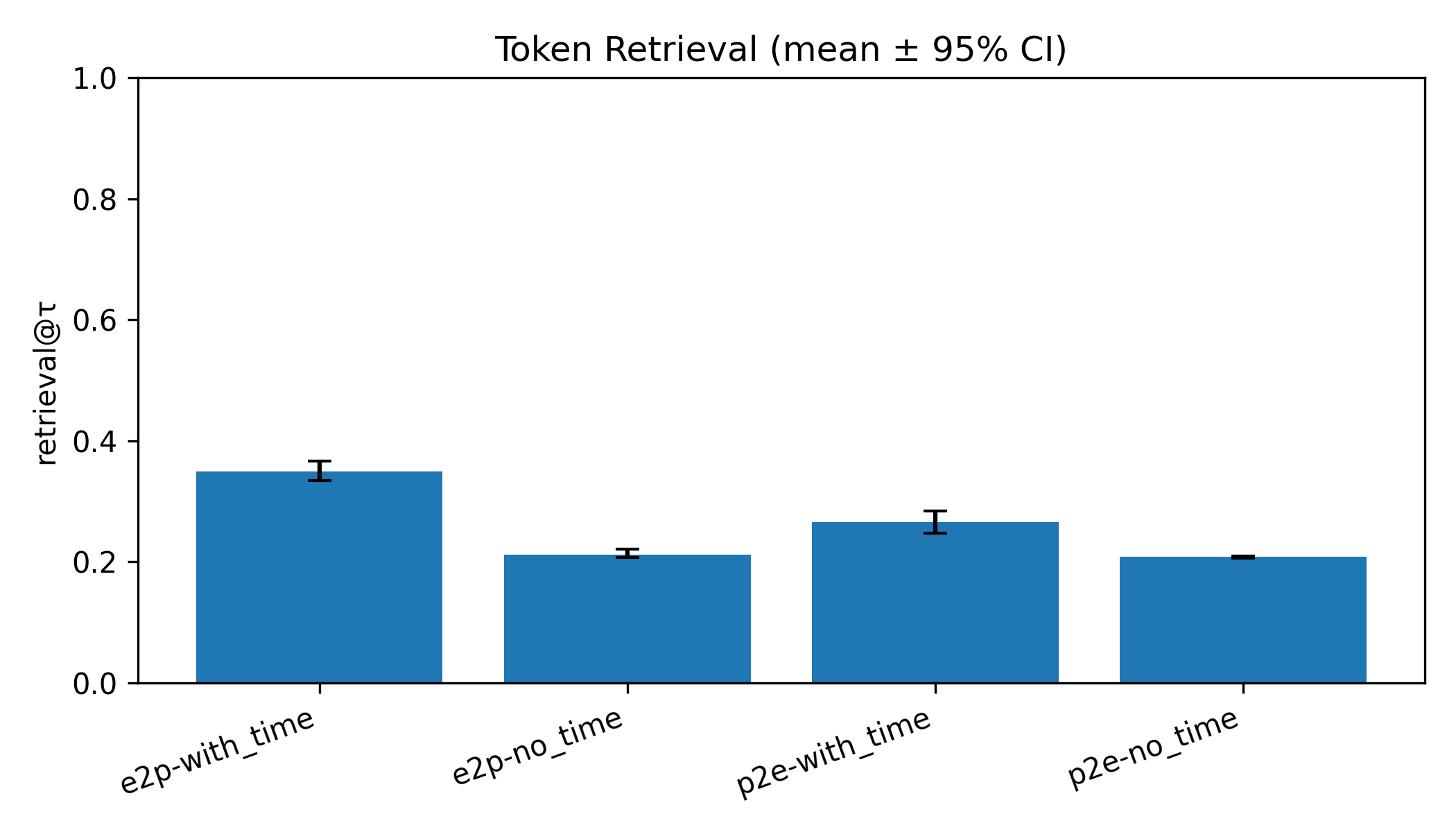}
    \caption{Cross-modal token-retrieval accuracy within tolerance $\tau$ seconds for EEG$\rightarrow$Phys and Phys$\rightarrow$EEG. Bars compare \emph{WITH time (CTAF)} to \emph{NO time} ablation; error bars show 95\% bootstrap CIs across participants.}
    \label{fig:retr-bars}
\end{figure}

\textbf{Token-level retrieval.} Figure \ref{fig:retr-bars}  reports cross-modal token retrieval within a tolerance $\tau$ seconds (EEG→physiology and physiology→EEG). Token retrieval is above chance and asymmetric in the expected direction—EEG→phys $0.350$ and phys→EEG $0.265$ at $\tau$ = $1$ s—both exceeding the no-time setting ($0.212$ and $0.208$). These rates increase substantially with time encodings, directly demonstrating learned cross-temporal correspondence. Interestingly, EEG→physiology and physiology→EEG rates are higher with time features than without, as summarized in Table \ref{tab:alignment_all}. Panel A and visualized in Figure 3. The asymmetry favoring EEG→physiology aligns with expected autonomic lags and further indicates that CTAF captures plausible cross-temporal correspondences.

Together, these results support the core claim that CTAF achieves robust, subject-generalizable cross-modal alignment.

\section{Discussion and Conclusion}

CTAF learns cross-temporal correspondences rather than assuming synchrony, estimating soft lag distributions between EEG and peripheral signals to construct contrastive positives, stabilize representations, and produce a shared clip embedding. In sensitivity analyses, we observed that the learned attention over lags can be broad for some held-out participants, consistent with consumer-grade EEG and window-level labels. We also found retrieval depends on the tolerance $\tau$ and on time encodings by tightening $\tau$ from 1.0 s to 0.5 s reduces retrieval via near-misses, and time-aware models consistently outperform no-time variants. These behaviors suggest straightforward extensions, such as sharper priors or temperature schedules to improve identifiability, principled $\tau$ selection, and alternative time encodings. 
\vspace{1pt}

CTAF provides a label-efficient, mask-safe mechanism for time-aware fusion that directly optimizes and evaluates cross-modal alignment. On the K-EmoCon dataset, it yields stronger alignment than time-agnostic variants and compares favorably to a supervised HyperFuseNet baseline while targeting a different objective. We see CTAF as a useful building block for brain–body representation learning, which handles asynchrony, scales to weak labels, and offers diagnostics aligned with the fusion goal. While our evaluation is limited to one dataset, in future work, we will validate generalizability on broader benchmarks and address the computational cost of attention mechanisms.

\bibliographystyle{abbrvnat}
\bibliography{neurips_2025}

@article{Park_2020,
   title={K-EmoCon, a multimodal sensor dataset for continuous emotion recognition in naturalistic conversations},
   volume={7},
   ISSN={2052-4463},
   url={http://dx.doi.org/10.1038/s41597-020-00630-y},
   DOI={10.1038/s41597-020-00630-y},
   number={1},
   journal={Scientific Data},
   publisher={Springer Science and Business Media LLC},
   author={Park, Cheul Young and Cha, Narae and Kang, Soowon and Kim, Auk and Khandoker, Ahsan Habib and Hadjileontiadis, Leontios and Oh, Alice and Jeong, Yong and Lee, Uichin},
   year={2020},
   month=sep 
}

@misc{eldele2021timeseriesrepresentationlearningtemporal,
      title={Time-Series Representation Learning via Temporal and Contextual Contrasting}, 
      author={Emadeldeen Eldele and Mohamed Ragab and Zhenghua Chen and Min Wu and Chee Keong Kwoh and Xiaoli Li and Cuntai Guan},
      year={2021},
      eprint={2106.14112},
      archivePrefix={arXiv},
      primaryClass={cs.LG},
      url={https://arxiv.org/abs/2106.14112}, 
}

@misc{yue2022ts2vecuniversalrepresentationtime,
      title={TS2Vec: Towards Universal Representation of Time Series}, 
      author={Zhihan Yue and Yujing Wang and Juanyong Duan and Tianmeng Yang and Congrui Huang and Yunhai Tong and Bixiong Xu},
      year={2022},
      eprint={2106.10466},
      archivePrefix={arXiv},
      primaryClass={cs.LG},
      url={https://arxiv.org/abs/2106.10466}, 
}

@ARTICLE{8269806,
  author={Baltrušaitis, Tadas and Ahuja, Chaitanya and Morency, Louis-Philippe},
  journal={IEEE Transactions on Pattern Analysis and Machine Intelligence}, 
  title={Multimodal Machine Learning: A Survey and Taxonomy}, 
  year={2019},
  volume={41},
  number={2},
  pages={423-443},
  doi={10.1109/TPAMI.2018.2798607}}

@misc{jiang2025multimodaltimeseriesanalysis,
      title={Multi-modal Time Series Analysis: A Tutorial and Survey}, 
      author={Yushan Jiang and Kanghui Ning and Zijie Pan and Xuyang Shen and Jingchao Ni and Wenchao Yu and Anderson Schneider and Haifeng Chen and Yuriy Nevmyvaka and Dongjin Song},
      year={2025},
      eprint={2503.13709},
      archivePrefix={arXiv},
      primaryClass={cs.LG},
      url={https://arxiv.org/abs/2503.13709}, 
}

@article{16d2153c76974234a7af4ed017b3bb68,
title = "DEAP: A Database for Emotion Analysis Using Physiological Signals",
author = "Sander Koelstra and C. M{\"u}hl and Mohammad Soleymani and Lee, \{Jung Seok\} and Ashkan Yazdani and Touradj Ebrahimi and Thierry Pun and Antinus Nijholt and Ioannis Patras",
note = "eemcs-eprint-21368 ",
year = "2012",
doi = "10.1109/T-AFFC.2011.15",
language = "Undefined",
volume = "3",
pages = "18--31",
journal = "IEEE transactions on affective computing",
issn = "1949-3045",
publisher = "IEEE",
number = "1",
}

@misc{cuturi2018softdtwdifferentiablelossfunction,
      title={Soft-DTW: a Differentiable Loss Function for Time-Series}, 
      author={Marco Cuturi and Mathieu Blondel},
      year={2018},
      eprint={1703.01541},
      archivePrefix={arXiv},
      primaryClass={stat.ML},
      url={https://arxiv.org/abs/1703.01541}, 
}

@misc{sermanet2018timecontrastivenetworksselfsupervisedlearning,
      title={Time-Contrastive Networks: Self-Supervised Learning from Video}, 
      author={Pierre Sermanet and Corey Lynch and Yevgen Chebotar and Jasmine Hsu and Eric Jang and Stefan Schaal and Sergey Levine},
      year={2018},
      eprint={1704.06888},
      archivePrefix={arXiv},
      primaryClass={cs.CV},
      url={https://arxiv.org/abs/1704.06888}, 
}

@misc{wang2019learningcorrespondencecycleconsistencytime,
      title={Learning Correspondence from the Cycle-Consistency of Time}, 
      author={Xiaolong Wang and Allan Jabri and Alexei A. Efros},
      year={2019},
      eprint={1903.07593},
      archivePrefix={arXiv},
      primaryClass={cs.CV},
      url={https://arxiv.org/abs/1903.07593}, 
}

@misc{radford2021learningtransferablevisualmodels,
      title={Learning Transferable Visual Models From Natural Language Supervision}, 
      author={Alec Radford and Jong Wook Kim and Chris Hallacy and Aditya Ramesh and Gabriel Goh and Sandhini Agarwal and Girish Sastry and Amanda Askell and Pamela Mishkin and Jack Clark and Gretchen Krueger and Ilya Sutskever},
      year={2021},
      eprint={2103.00020},
      archivePrefix={arXiv},
      primaryClass={cs.CV},
      url={https://arxiv.org/abs/2103.00020}, 
}

@misc{chen2020simpleframeworkcontrastivelearning,
      title={A Simple Framework for Contrastive Learning of Visual Representations}, 
      author={Ting Chen and Simon Kornblith and Mohammad Norouzi and Geoffrey Hinton},
      year={2020},
      eprint={2002.05709},
      archivePrefix={arXiv},
      primaryClass={cs.LG},
      url={https://arxiv.org/abs/2002.05709}, 
}

@misc{bardes2022vicregvarianceinvariancecovarianceregularizationselfsupervised,
      title={VICReg: Variance-Invariance-Covariance Regularization for Self-Supervised Learning}, 
      author={Adrien Bardes and Jean Ponce and Yann LeCun},
      year={2022},
      eprint={2105.04906},
      archivePrefix={arXiv},
      primaryClass={cs.CV},
      url={https://arxiv.org/abs/2105.04906}, 
}

@misc{zbontar2021barlowtwinsselfsupervisedlearning,
      title={Barlow Twins: Self-Supervised Learning via Redundancy Reduction}, 
      author={Jure Zbontar and Li Jing and Ishan Misra and Yann LeCun and Stéphane Deny},
      year={2021},
      eprint={2103.03230},
      archivePrefix={arXiv},
      primaryClass={cs.CV},
      url={https://arxiv.org/abs/2103.03230}, 
}

@INPROCEEDINGS{10193329,
  author={Lopez, Eleonora and Chiarantano, Eleonora and Grassucci, Eleonora and Comminiello, Danilo},
  booktitle={2023 IEEE International Conference on Acoustics, Speech, and Signal Processing Workshops (ICASSPW)}, 
  title={Hypercomplex Multimodal Emotion Recognition from EEG and Peripheral Physiological Signals}, 
  year={2023},
  volume={},
  number={},
  pages={1-5},
  doi={10.1109/ICASSPW59220.2023.10193329}}

@misc{lopez2024hierarchicalhypercomplexnetworkmultimodal,
      title={Hierarchical Hypercomplex Network for Multimodal Emotion Recognition}, 
      author={Eleonora Lopez and Aurelio Uncini and Danilo Comminiello},
      year={2024},
      eprint={2409.09194},
      archivePrefix={arXiv},
      primaryClass={cs.LG},
      url={https://arxiv.org/abs/2409.09194}, 
}

@misc{lopez2024phemonetmultimodalnetworkphysiological,
      title={PHemoNet: A Multimodal Network for Physiological Signals}, 
      author={Eleonora Lopez and Aurelio Uncini and Danilo Comminiello},
      year={2024},
      eprint={2410.00010},
      archivePrefix={arXiv},
      primaryClass={eess.SP},
      url={https://arxiv.org/abs/2410.00010}, 
}

@article{article,
author = {Kreibig, Sylvia},
year = {2010},
month = {04},
pages = {394-421},
title = {Autonomic Nervous System Activity in Emotion: A Review},
volume = {84},
journal = {Biological psychology},
doi = {10.1016/j.biopsycho.2010.03.010}
}

@ARTICLE{8554112,
  author={Miranda-Correa, Juan Abdon and Abadi, Mojtaba Khomami and Sebe, Nicu and Patras, Ioannis},
  journal={IEEE Transactions on Affective Computing}, 
  title={AMIGOS: A Dataset for Affect, Personality and Mood Research on Individuals and Groups}, 
  year={2021},
  volume={12},
  number={2},
  pages={479-493},
  doi={10.1109/TAFFC.2018.2884461}}

@inproceedings{10.1145/3242969.3242985,
author = {Schmidt, Philip and Reiss, Attila and Duerichen, Robert and Marberger, Claus and Van Laerhoven, Kristof},
title = {Introducing WESAD, a Multimodal Dataset for Wearable Stress and Affect Detection},
year = {2018},
isbn = {9781450356923},
publisher = {Association for Computing Machinery},
address = {New York, NY, USA},
url = {https://doi.org/10.1145/3242969.3242985},
doi = {10.1145/3242969.3242985},
booktitle = {Proceedings of the 20th ACM International Conference on Multimodal Interaction},
pages = {400–408},
numpages = {9},
location = {Boulder, CO, USA},
series = {ICMI '18}
}

@misc{oord2019representationlearningcontrastivepredictive,
      title={Representation Learning with Contrastive Predictive Coding}, 
      author={Aaron van den Oord and Yazhe Li and Oriol Vinyals},
      year={2019},
      eprint={1807.03748},
      archivePrefix={arXiv},
      primaryClass={cs.LG},
      url={https://arxiv.org/abs/1807.03748}, 
}

@misc{tonekaboni2021unsupervisedrepresentationlearningtime,
      title={Unsupervised Representation Learning for Time Series with Temporal Neighborhood Coding}, 
      author={Sana Tonekaboni and Danny Eytan and Anna Goldenberg},
      year={2021},
      eprint={2106.00750},
      archivePrefix={arXiv},
      primaryClass={cs.LG},
      url={https://arxiv.org/abs/2106.00750}, 
}

@InProceedings{pmlr-v28-andrew13,
  title = 	 {Deep Canonical Correlation Analysis},
  author = 	 {Andrew, Galen and Arora, Raman and Bilmes, Jeff and Livescu, Karen},
  booktitle = 	 {Proceedings of the 30th International Conference on Machine Learning},
  pages = 	 {1247--1255},
  year = 	 {2013},
  editor = 	 {Dasgupta, Sanjoy and McAllester, David},
  volume = 	 {28},
  number =       {3},
  series = 	 {Proceedings of Machine Learning Research},
  address = 	 {Atlanta, Georgia, USA},
  month = 	 {17--19 Jun},
  publisher =    {PMLR},
  url = 	 {https://proceedings.mlr.press/v28/andrew13.html}
}

@inproceedings{tsai-etal-2019-multimodal,
    title = "Multimodal Transformer for Unaligned Multimodal Language Sequences",
    author = "Tsai, Yao-Hung Hubert  and
      Bai, Shaojie  and
      Liang, Paul Pu  and
      Kolter, J. Zico  and
      Morency, Louis-Philippe  and
      Salakhutdinov, Ruslan",
    editor = "Korhonen, Anna  and
      Traum, David  and
      M{\`a}rquez, Llu{\'i}s",
    booktitle = "Proceedings of the 57th Annual Meeting of the Association for Computational Linguistics",
    month = jul,
    year = "2019",
    address = "Florence, Italy",
    publisher = "Association for Computational Linguistics",
    url = "https://aclanthology.org/P19-1656/",
    doi = "10.18653/v1/P19-1656",
    pages = "6558--6569"
}

@misc{dwibedi2019temporalcycleconsistencylearning,
      title={Temporal Cycle-Consistency Learning}, 
      author={Debidatta Dwibedi and Yusuf Aytar and Jonathan Tompson and Pierre Sermanet and Andrew Zisserman},
      year={2019},
      eprint={1904.07846},
      archivePrefix={arXiv},
      primaryClass={cs.CV},
      url={https://arxiv.org/abs/1904.07846}, 
}

@inproceedings{10.1145/3608164.3608185,
author = {Wang, Yingxi and Liang, Huizhi and Zhai, Bing},
title = {Temporal Neighborhood based Self-supervised Pre-training Model for Sleep Stages Classification},
year = {2023},
isbn = {9798400700385},
publisher = {Association for Computing Machinery},
address = {New York, NY, USA},
url = {https://doi.org/10.1145/3608164.3608185},
doi = {10.1145/3608164.3608185},
booktitle = {Proceedings of the 2023 15th International Conference on Bioinformatics and Biomedical Technology},
pages = {149–155},
numpages = {7},
location = {Xi'an, China},
series = {ICBBT '23}
}

@misc{guo2023emotionrecognitionbasedmultimodal,
      title={Emotion recognition based on multi-modal electrophysiology multi-head attention Contrastive Learning}, 
      author={Yunfei Guo and Tao Zhang and Wu Huang},
      year={2023},
      eprint={2308.01919},
      archivePrefix={arXiv},
      primaryClass={cs.MM},
      url={https://arxiv.org/abs/2308.01919}, 
}

@article{BENEDEK201080,
title = {A continuous measure of phasic electrodermal activity},
journal = {Journal of Neuroscience Methods},
volume = {190},
number = {1},
pages = {80-91},
year = {2010},
issn = {0165-0270},
doi = {https://doi.org/10.1016/j.jneumeth.2010.04.028},
url = {https://www.sciencedirect.com/science/article/pii/S0165027010002335},
author = {Mathias Benedek and Christian Kaernbach}
}

@article{THAYER2000201,
title = {A model of neurovisceral integration in emotion regulation and dysregulation},
journal = {Journal of Affective Disorders},
volume = {61},
number = {3},
pages = {201-216},
year = {2000},
note = {Arousal in Anxiety},
issn = {0165-0327},
doi = {https://doi.org/10.1016/S0165-0327(00)00338-4},
url = {https://www.sciencedirect.com/science/article/pii/S0165032700003384},
author = {Julian F Thayer and Richard D Lane}
}

@misc{aristimunha2025eegfoundationchallengecrosstask,
      title={EEG Foundation Challenge: From Cross-Task to Cross-Subject EEG Decoding}, 
      author={Bruno Aristimunha and Dung Truong and Pierre Guetschel and Seyed Yahya Shirazi and Isabelle Guyon and Alexandre R. Franco and Michael P. Milham and Aviv Dotan and Scott Makeig and Alexandre Gramfort and Jean-Remi King and Marie-Constance Corsi and Pedro A. Valdés-Sosa and Amit Majumdar and Alan Evans and Terrence J Sejnowski and Oren Shriki and Sylvain Chevallier and Arnaud Delorme},
      year={2025},
      eprint={2506.19141},
      archivePrefix={arXiv},
      primaryClass={eess.SP},
      url={https://arxiv.org/abs/2506.19141}, 
}

@misc{grill2020bootstraplatentnewapproach,
      title={Bootstrap your own latent: A new approach to self-supervised Learning}, 
      author={Jean-Bastien Grill and Florian Strub and Florent Altché and Corentin Tallec and Pierre H. Richemond and Elena Buchatskaya and Carl Doersch and Bernardo Avila Pires and Zhaohan Daniel Guo and Mohammad Gheshlaghi Azar and Bilal Piot and Koray Kavukcuoglu and Rémi Munos and Michal Valko},
      year={2020},
      eprint={2006.07733},
      archivePrefix={arXiv},
      primaryClass={cs.LG},
      url={https://arxiv.org/abs/2006.07733}, 
}

@misc{kiyasseh2021clocscontrastivelearningcardiac,
      title={CLOCS: Contrastive Learning of Cardiac Signals Across Space, Time, and Patients}, 
      author={Dani Kiyasseh and Tingting Zhu and David A. Clifton},
      year={2021},
      eprint={2005.13249},
      archivePrefix={arXiv},
      primaryClass={cs.LG},
      url={https://arxiv.org/abs/2005.13249}, 
}

@misc{vaswani2023attentionneed,
      title={Attention Is All You Need}, 
      author={Ashish Vaswani and Noam Shazeer and Niki Parmar and Jakob Uszkoreit and Llion Jones and Aidan N. Gomez and Lukasz Kaiser and Illia Polosukhin},
      year={2023},
      eprint={1706.03762},
      archivePrefix={arXiv},
      primaryClass={cs.CL},
      url={https://arxiv.org/abs/1706.03762}, 
}
\end{document}